\documentclass{article}

\usepackage{PRIMEarxiv}

\usepackage[utf8]{inputenc} 
\usepackage[T1]{fontenc}    
\usepackage{hyperref}       
\usepackage{url}            
\usepackage{booktabs}       
\usepackage{amsfonts}       
\usepackage{nicefrac}       
\usepackage{microtype}      
\usepackage{lipsum}
\usepackage{natbib}
\usepackage{fancyhdr}       
\usepackage{graphicx}       
\graphicspath{{media/}}     
\usepackage{algorithmic}
\usepackage{algorithm}
\usepackage{amsmath}
\usepackage{mathtools}
\usepackage{relsize}
\usepackage{enumitem}
\usepackage[table,xcdraw]{xcolor}
\usepackage{colortbl} 
\usepackage[normalem]{ulem}
\useunder{\uline}{\ul}{}
\usepackage{multirow}

\title{ACPO: Adaptive Curriculum Policy Optimization for Aligning Vision-Language Models in Complex Reasoning
}

\author{
  Yunhao Wang, Ziting Li*, Shuai Chen*, Tao Liu, Chao Song, Junjie Jiang, Jian Zhu, Peng Gao, Bin Qin \\
  Xiaomi Inc., Beijing, China \\
  \texttt{\{wangyunhao,songchao3,liutao39,jiangjunjie,zhujian5,gaopeng11,qinbin\}@xiaomi.com} \\
   \And
}

\begin{document}
\maketitle

\begin{abstract}
Aligning large-scale vision-language models (VLMs) for complex reasoning via reinforcement learning is often hampered by the limitations of existing policy optimization algorithms, such as static training schedules and the rigid, uniform clipping mechanism in Proximal Policy Optimization (PPO). In this work, we introduce Adaptive Curriculum Policy Optimization (ACPO), a novel framework that addresses these challenges through a dual-component adaptive learning strategy. First, ACPO employs a dynamic curriculum that orchestrates a principled transition from a stable, near on-policy exploration phase to an efficient, off-policy exploitation phase by progressively increasing sample reuse. Second, we propose an Advantage-Aware Adaptive Clipping (AAAC) mechanism that replaces the fixed clipping hyperparameter with dynamic, sample-wise bounds modulated by the normalized advantage of each token. This allows for more granular and robust policy updates, enabling larger gradients for high-potential samples while safeguarding against destructive ones. We conduct extensive experiments on a suite of challenging multimodal reasoning benchmarks, including MathVista, LogicVista, and MMMU-Pro. Results demonstrate that ACPO consistently outperforms strong baselines such as DAPO and PAPO, achieving state-of-the-art performance, accelerated convergence, and superior training stability.
\end{abstract}

\section{Introduction}
Large-scale vision-language models (VLMs) like CLIP~\cite{radford2021learning}, Flamingo~\cite{alayrac2022flamingo}, and GPT-4V~\cite{openai2023gpt4} have achieved remarkable progress in multimodal reasoning, from solving scientific diagram problems to tackling complex visual question-answering tasks. However, to truly excel at intricate, domain-specific reasoning, these models require a crucial final step: alignment ~\cite{kirk2023understanding}. Recent advancements have adapted reinforcement learning from human feedback (RLHF) to this domain, with methods such as GRPO~\cite{deepseek}, DAPO~\cite{dapo}, and PAPO~\cite{huang2024papo} demonstrating impressive results. While these approaches have significantly improved sample efficiency, they often rely on static training schedules and a rigid, one-size-fits-all clipping mechanism inherited from Proximal Policy Optimization (PPO)~\cite{schulman2017ppo}. This fixed clipping can either suppress beneficial policy updates for high-potential tokens or fail to constrain destructive ones from noisy signals, leading to training instability and suboptimal performance.

To overcome these limitations, we introduce \textbf{Adaptive Curriculum Policy Optimization (ACPO)}, a novel RL framework that adapts its learning strategy dynamically to the model's evolving capabilities. Unlike prior methods that use fixed hyperparameters and schedules, ACPO employs a dual-component adaptive learning strategy to enhance both training stability and sample efficiency.

First, we propose a \textbf{dynamic curriculum policy} that orchestrates a principled transition between learning phases. Instead of a static schedule, ACPO starts in a stable, near-on-policy exploration phase with frequent data refreshes and short reuse windows, ensuring stable gradients and a robust policy foundation. As training progresses, the curriculum automatically shifts to an efficient, off-policy exploitation phase, where sample reuse is progressively increased. This allows the model to intensively fine-tune its policy on high-quality data, accelerating convergence without risking overfitting or catastrophic forgetting.

Second, we introduce an \textbf{Advantage-Aware Adaptive Clipping (AAAC)} mechanism that fundamentally refines PPO's update rule. Traditional PPO uses a fixed clipping threshold that applies uniformly to all samples. In contrast, our AAAC mechanism replaces this with dynamic, sample-wise bounds that are modulated by the normalized advantage of each token. This allows for a fine-grained allocation of the gradient budget: high-advantage samples with strong learning signals are granted a wider clipping range, enabling more aggressive and precise updates, while low- or negative-advantage samples are conservatively constrained to prevent instability. This dynamic control over the optimization landscape improves both learning efficiency and policy robustness.

We conduct extensive experiments on a suite of challenging multimodal reasoning benchmarks, including MathVista~\cite{luo2023mathvista}, LogicVista~\cite{wang2024logicvista}, DynaMath~\cite{zhang2024dynamath}, and MMMU-Pro~\cite{yu2023mmmu}. Our results consistently demonstrate that ACPO outperforms state-of-the-art baselines, including DAPO and PAPO, achieving superior performance, faster convergence, and enhanced training stability across all tasks.

Our contributions can be summarized as follows:
\begin{itemize}[left = 0pt]
    \item We propose a novel dynamic curriculum framework that automatically balances on-policy exploration and off-policy exploitation, allowing the training process to evolve in synchrony with the model's capabilities.
    \item The introduction of AAAC provides a new mechanism that replaces PPO's fixed clipping threshold with dynamic, advantage-modulated bounds, enabling more effective and robust policy updates.
    \item Through extensive evaluation, we demonstrate that ACPO achieves sota performance and accelerated convergence on a comprehensive suite of complex multimodal reasoning benchmarks.
\end{itemize}

\section{Related Work}

\subsection{Reinforcement Learning from Human Feedback}
RLHF has become the dominant paradigm for aligning large language models with human preferences~\cite{ouyang2022training, kaufmann2023survey}. Early approaches typically employ Proximal Policy Optimization (PPO)~\cite {schulman2017ppo} for training stabilization, with a static clipping mechanism adopted to constrain policy updates. While effective, PPO's uniform clipping can lead to suboptimal updates, suppressing high-advantage signals or failing to constrain harmful updates, which may cause instability or entropy collapse.

Building upon this, subsequent algorithms have sought to refine the optimization process. GRPO introduced a group-based reward formulation that aggregates responses per prompt and computes a shared advantage signal across all generated outputs, which improves training stability by reducing variance in reward estimation. It also employs a token-level Kullback-Leibler (KL) penalty to prevent excessive deviation from the reference policy at the sequence level, thereby mitigating mode collapse while preserving fine-grained control over generation. DAPO introduced several key improvements to enhance stability and sample efficiency. To counter entropy collapse, DAPO proposed the clip-higher strategy, which asymmetrically increases the upper clipping bound to encourage exploration. To address vanishing gradients for prompts with near-perfect or zero accuracy, it introduced a Dynamic Sampling mechanism to filter out these less informative instances. DAPO also incorporated token-level loss and a soft penalty for overlong responses to further stabilize training. 

 Furthermore, Macro-Action RLHF (MA-RLHF)~\cite{chai2024ma_rlhf} introduces macro actions, such as token sequences or higher-level language structures, to reduce credit assignment issues over long horizons and improve learning efficiency. Contrastive reward mechanisms~\cite{shen2024contrastive} reduce uncertainty in reward models and encourage improvement beyond baseline performance, mitigating variance issues in PPO. Personalized RLHF approaches~\cite{poddar2024personalized} capture diverse user preferences using variational methods, enabling personalized reward modeling and better performance across different user populations. Unsupervised RLHF~\cite{solway2024unsupervised} leverages signals derived automatically from data to provide negative guidance, enabling fine-grained model adjustment without the need for additional human feedback. Reward ensemble methods~\cite{zhang2024ensemble} combine multiple reward models to enhance prediction accuracy, addressing errors caused by limited training data in conventional RLHF.

While these efforts have broadened the horizons of reward modeling, policy updates, and personalized alignment, achieving robust training stability and high sample efficiency remains a central challenge. To this end, we introduce ACPO, a method that incorporates a dynamic curriculum and Advantage-Aware Adaptive Clipping to directly tackle these issues.

\subsection{Curriculum Learning for Reinforcement Learning}
Curriculum learning (CL) enhances training by organizing tasks in a progressive sequence, starting with simpler ones and gradually increasing complexity. This approach aims to improve both learning efficiency and generalization~\cite{bengio2009curriculum}. In RL, CL techniques include methods like task sorting by difficulty~\cite{wang2019paired,justesen2018illuminating}, teacher-student models that adaptively select tasks based on the learner's progress~\cite{portelas2020teacher}, and self-play strategies that create curricula through agent competition~\cite{sukhbaatar2017intrinsic}.

Although CL has been extensively explored in traditional RL, its use in RLHF for LLMs remains limited. Current approaches typically rely on staged training with predefined difficulty levels~\cite{wen2025light,luo2025deepscaler,song2025fastcurl} or online filtering techniques that sample and discard data until that rewards fall within a specific range~\cite{bae2025online,yu2025dapo}. However, these methods often lack adaptability due to dynamic difficulty levels in each batch of the training data.

In contrast, our framework actively guides the learning trajectory by fully considering the evolving nature of the training process. It employs a dual-component mechanism: a dynamic frequency control scheduler that orchestrates the transition from stable on-policy updates to efficient off-policy sample reuse, and a course-aware sample screening process that progressively increases the difficulty of training data. This structured approach ensures the model first masters foundational knowledge before focusing on more challenging examples, leading to more robust and efficient convergence.

\section{Method}

\begin{figure}[!t]
\centering
\includegraphics[width=0.95\linewidth]{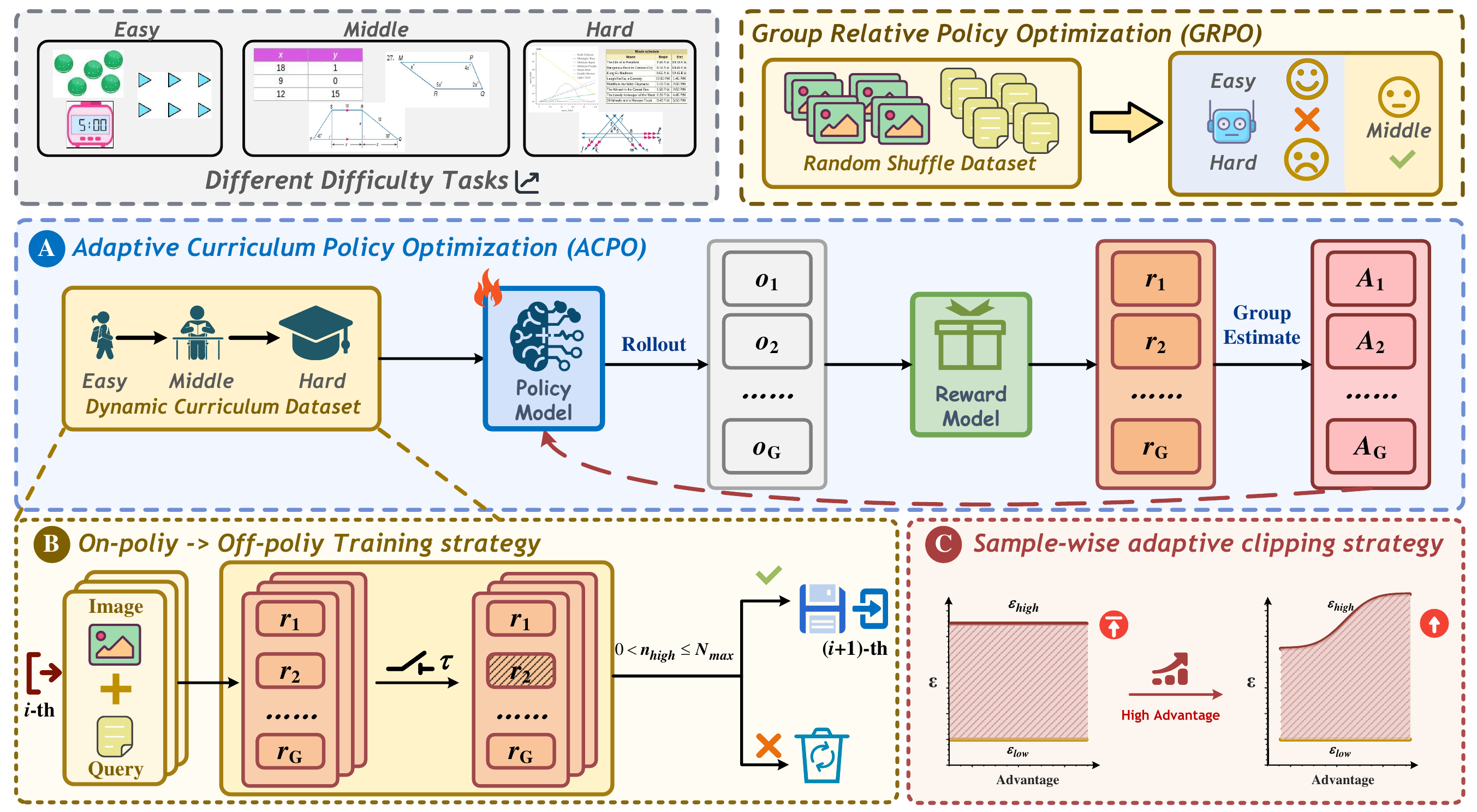}
\caption{Overview of ACPO. Unlike GRPO, ACPO removes the KL divergence constraint. Module B introduces dynamic curriculum sampling, where the $i-th$ iteration selects moderately difficult samples based on threshold $\tau$ and $N_{max}$, which then proceed to $(i+1)-th$ iteration . Module C adds advantage-based clipping, enabling safer, more effective updates for high-advantage samples.}
\label{fig1}
\end{figure}

\subsection{Overview}

Aligning VLMs for complex reasoning via RLHF has become a predominant paradigm. While recent algorithms like GRPO and DAPO have achieved significant gains in sample efficiency and performance, they are often limited by static training schedules and a fixed clipping threshold in PPO. This rigid, one-size-fits-all mechanism can be suboptimal, either suppressing beneficial policy updates or failing to prevent destructive ones, which leads to training instability and limits the model's potential. To overcome these challenges, we introduce ACPO, a novel framework illustrated in Fig. \ref{fig1}. Our approach features two key innovations: a dynamic curriculum that intelligently transitions training from a stable on-policy to an efficient off-policy regime (Fig. \ref{fig1}(B)), and a novel sample-wise adaptive clipping mechanism that modulates optimization bounds on a per-sample basis according to its advantage (Fig. \ref{fig1}(C)). This dual approach significantly enhances training stability and convergence efficiency, leading to state-of-the-art performance.

\subsubsection{Strategic Gating Sampling}
To enhance training stability and focus the model on high-quality signals, ACPO first employs a strategic sample gating mechanism. At each training step $t$, for a candidate batch of queries $\mathcal{B}_t = \{q_j\}_{j=1}^M$, we generate responses using the reference policy $\pi_{\theta_{\text{old}}}$. This batch is then filtered to produce a high-quality subset, $\mathcal{B}_{\text{valid}}$, based on reward and diversity criteria:
\begin{equation}
\label{eq:gating}
\mathcal{B}_{\text{valid}} = \left\{ q \in \mathcal{B}_t \mid 0 < \sum_{i=1}^G \mathbb{I}(R(o_i) > \tau) \leq N_{\text{max}} \right\}
\end{equation}
where $\tau$ is the minimum reward threshold and $N_{\text{max}}$ is the maximum number of high-reward responses per query, which encourages diversity. Only queries that elicit a sufficient number of high-reward responses are retained in $\mathcal{B}_{\text{valid}}$ for the subsequent optimization phase.

\subsubsection{On-Policy to Off-Policy Phase Transition}
After identifying the valid samples, ACPO uses its dynamic curriculum to manage the stability-efficiency trade-off, governed by the adaptive reuse count $K(t)$.

The GRPO objective's expectation is taken over the original, unfiltered batch $\mathcal{B}_t$. The crucial link to the gating mechanism is established by incorporating an indicator function, $\mathbb{I}(q \in \mathcal{B}_{\text{valid}})$, which effectively masks out the loss for any sample that did not meet the gating criteria:
\begin{equation}
\label{eq:objective}
J_{\text{GRPO}}(\theta) =
\mathbb{E}_{q \sim \mathcal{B}_{\text{valid}},\, \{o_i\} \sim \pi_{\theta_{\text{old}}}} \left[
    \frac{1}{G} \sum_{i=1}^G \frac{1}{|o_i|} \sum_{t=1}^{|o_i|}
    \Big( \min(r_{i,t}(\theta)\hat{A}_{i,t},\, c_{i,t}) - \beta\, D_{\text{KL}}(\pi_\theta \,\|\, \pi_{\text{ref}}) \Big)
\right],
\end{equation}

where $r_{i,t}(\theta) = \tfrac{\pi_\theta(o_{i,t}\mid q, o_{i,<t})}{\pi_{\theta_{\text{old}}}(o_{i,t}\mid q, o_{i,<t})}$ is the probability ratio, $\hat{A}_{i,t}$ is the advantage estimate, and 
$c_{i,t} = \text{clip}(r_{i,t}(\theta),\, 1-\epsilon,\, 1+\epsilon)\,\hat{A}_{i,t}$ denotes the clipped objective term. Importantly, the expectation in Eq.~\ref{eq:objective} is taken over queries $q$ drawn from the gated batch $\mathcal{B}_{\text{valid}}$ (see Eq.~\ref{eq:gating}), ensuring that only high-quality samples contribute to the policy gradient update.

Instead of performing a fixed number of updates, ACPO performs $K(t)$ optimization steps using the objective in Eq. ~\ref{eq:objective}, where $K(t)$ adapts with training progress:
\begin{equation}
\label{eq:schedule}
    K(t) = \max\left(1, \left\lceil \frac{N\cdot t}{T} \right\rceil\right)
\end{equation}
where $N$ is the maximum reuse count, $t$ is the current training step, and $T$ is the total duration. This curriculum creates a principled transition through three distinct phases:
\begin{itemize}[left = 0pt]
    \item \textbf{On-policy Exploration Phase ($t \ll T$)}: When $K(t) \approx 1$, the model prioritizes stable learning on fresh, high-quality data to build a robust policy foundation.
    \item \textbf{Balanced Transition Phase}: As $K(t)$ grows linearly, the strategy gradually anneals towards off-policy learning, increasing sample reuse as the policy stabilizes.
    \item \textbf{Off-policy Exploitation Phase ($t \rightarrow T$)}: When $K(t) \rightarrow N$, the model intensively fine-tunes its policy on the most valuable gated samples, maximizing data utility to accelerate final convergence.
\end{itemize}

\subsection{Advantage-Aware Adaptive Clipping}
A primary limitation of the standard PPO algorithm is its reliance on a fixed clipping hyperparameter, $\epsilon$, which applies a uniform update constraint to all samples regardless of their learning potential. This can either stifle progress on high-quality samples or fail to prevent destructive updates from noisy ones.

To overcome this, ACPO introduces an \textbf{Advantage-Aware Adaptive Clipping} mechanism. Instead of a static bound, the upper clipping range is dynamically modulated by the magnitude of the sample's advantage, allowing for a more granular and intelligent policy update. The ACPO objective is formulated as:

\begin{equation}
\label{eq:ACPO_objective}
J_{\text{ACPO}}(\theta) =
\mathbb{E}_{q \sim \mathcal{B}_{\text{valid}}, \{o_i\} \sim \pi_{\theta_{\text{old}}}(\cdot|q)}
\Bigg[
\frac{1}{\sum_{i=1}^G |o_i|} \sum_{i=1}^G \sum_{t=1}^{|o_i|}
\min\big(r_{i,t}(\theta)\hat{A}_{i,t},\, c_{i,t}\big)
\Bigg],
\end{equation}

where $r_{i,t}(\theta)$ is defined as $\frac{\pi_\theta(o_{i,t}|q)}{\pi_{\theta_{\text{old}}}(o_{i,t}|q)}$, and 
$c_{i,t} = \text{clip}\Big(r_{i,t}(\theta),\, 1 - \epsilon_{\text{low}},\, 1 + \epsilon_{\text{high}}(\hat{A}_{i,t})\Big) \hat{A}_{i,t}$.

The key innovation lies in the upper clipping bound, $\epsilon_{\text{high}}$, which is no longer a fixed value but a function of the token-level advantage $\hat{A}_{i,t}$:
\begin{equation}
\label{eq:epsilon_high}
\epsilon_{\text{high}}(\hat{A}_{i,t}) = \epsilon_{\text{high}}^0 + \delta \cdot \tilde{A}_{i,t}
\end{equation}
where $\epsilon_{\text{high}}^0$ is a baseline clipping value and $\delta$ is a scaling factor controlling the sensitivity to the advantage. The term $\tilde{A}_{i,t}$ represents the normalized advantage, which is transformed from an unbounded range to $[0, 1]$ using the error function (\textbf{erf}):
\begin{equation}
\label{eq:advantage_norm}
\tilde{A}_{i,t} = \frac{1}{2}\left(1 + \textbf{erf}\left(\frac{\hat{A}_{i,t}}{\sqrt{2}\sigma_A}\right)\right)
\end{equation}
where $\sigma_A$ is the standard deviation of the advantages in the batch, used for scaling. This formulation establishes a fine-grained, sample-wise control over the optimization landscape. High-advantage samples are rewarded with a significantly wider clipping range, enabling larger and more confident policy updates that capitalize on strong learning signals. Conversely, low- or negative-advantage samples are met with a conservative bound that shields the policy from noisy or potentially destructive gradients. In essence, this mechanism allows ACPO to dynamically allocate its gradient budget—accelerating convergence by exploiting high-potential updates while preserving the stability crucial for complex reasoning tasks.

The entire process, which integrates strategic data gating with an adaptive update curriculum, is summarized in Alg.~\ref{alg:ACPO}. The pseudocode outlines the complete training loop, from data sampling and filtering to the dynamically scheduled, advantage-aware policy updates.

\begin{algorithm}[htbp]
\caption{Dynamic Curriculum Policy Optimization (ACPO)}
\label{alg:ACPO}
\begin{algorithmic}[1]
    \STATE \textbf{Input}: Initial policy $\pi_{\theta_{\text{init}}}$, reward model $r_\phi$, prompt dataset $\mathcal{D}$.
    \STATE \textbf{Hyperparameters}: Max reuse count $N$, outer iterations $I$, training steps per iteration $T$, batch size $M$, clipping baseline $\epsilon_{\text{high}}^0$, sensitivity $\delta$, reward threshold $\tau$, diversity count $N_{\text{max}}$.
    \STATE \textbf{Output}: Optimized policy $\pi_\theta^*$.
    
    \STATE Initialize policy $\pi_\theta \leftarrow \pi_{\theta_{\text{init}}}$.
    \FOR{iteration $i = 1$ to $I$}
        \STATE Set reference policy for KL penalty $\pi_{\text{ref}} \leftarrow \pi_\theta$.
        \FOR{training step $t = 1$ to $T$}
            \STATE Sample a batch of prompts $\mathcal{B}_t = \{q_j\}_{j=1}^M \sim \mathcal{D}$.
            \STATE Set old policy for sampling $\pi_{\theta_{\text{old}}} \leftarrow \pi_\theta$.
            \STATE Generate responses $\{o_i\}_{i=1}^G \sim \pi_{\theta_{\text{old}}}(\cdot|q)$ for each $q \in \mathcal{B}_t$.
            \STATE Construct the valid batch $\mathcal{B}_{\text{valid}} \subseteq \mathcal{B}_t$ using the gating criteria in Eq. ~\ref{eq:gating}.
            \STATE Compute rewards $R(o_i)$ for all responses using $r_\phi$.
            \STATE Compute advantages $\hat{A}_{i,t}$ for each token in all responses.
            
            \STATE \COMMENT{Begin adaptive update phase}
            \STATE Determine curriculum reuse count $K(t) \leftarrow \max(1, \lceil N \cdot t / T \rceil)$ using Eq. ~\ref{eq:schedule}.
            \FOR{update epoch $k = 1$ to $K(t)$}
                \STATE Compute loss $L(\theta)$ on batch $\mathcal{B}_t$ using the full ACPO objective $J_{\text{ACPO}}$ from Eq. ~\ref{eq:objective}.
                \STATE \COMMENT{The objective implicitly masks invalid samples and uses adaptive clipping.}
                \STATE Update policy parameters $\theta \leftarrow \text{optimizer\_step}(\theta, \nabla_\theta L(\theta))$.
            \ENDFOR
        \ENDFOR
    \ENDFOR
    \STATE \textbf{return} optimized policy $\pi_\theta^* \leftarrow \pi_\theta$.
\end{algorithmic}
\end{algorithm}

\section{Experiments}
\label{Experiments}

\subsection{Experimental Setup}

All experiments were conducted on four servers, each equipped with eight H20 GPUs. The training process utilized the DeepSpeed Zero2 ~\citep{deepspeed} configuration to optimize memory usage and efficiency. We based our models on the Qwen2.5-VL-3B ~\citep{qwen25vl}, training them on the ViRL39K dataset \cite{ViRL39K} with a learning rate of 1e-6 using direct reinforcement learning. Our proposed ACPO method was compared against standard baselines, including DAPO and PAPO. Since ACPO blends on-policy and off-policy approaches, we included both on-policy and off-policy DAPO baselines in our evaluation for a comprehensive comparison.

\subsubsection{Evaluation}

To comprehensively evaluate the effectiveness of our method, we conducted experiments on seven public benchmarks covering diverse reasoning domains. These include: Geometry3K ~\cite{Geometry3K}, MathVerse , MathVerse-V \cite{Mathverse}, and We-Math \cite{Wemath} for mathematical and geometric reasoning; MMMU-Pro \cite{Mmmupro} for multi-discipline multimodal reasoning; LogicVista \cite{Logicvista} for logical reasoning; and Counting \cite{Counting} for counting tasks. Evaluation was based on exact match between model predictions and ground-truth answers. We report the average accuracy@8 across all benchmarks with a reasoning temperature of 1.0. Datasets requiring free-form responses or those evaluated by LLM-based judges were excluded from this study.

\subsubsection{Main Results}

The superior performance of ACPO, as evidenced in Tab.~\ref{tab:main_results} and ~\ref{tab:main_results-2}, can be directly attributed to its two core methodological innovations: the dynamic on-policy to off-policy curriculum and the advantage-aware adaptive clipping mechanism. The consistent gains across both 3B and 7B scales—particularly in general reasoning tasks like MathVerse, Geo3k, and We-Math—reflect the effectiveness of ACPO’s strategic sample gating and phased training schedule. By initially operating in a stable on-policy regime, ACPO avoids the early-stage instability that often plagues off-policy methods, allowing the policy to establish a reliable foundation. As training progresses, the linear increase in reuse count $K(t)$ enables efficient exploitation of high-reward, gated samples, which explains the pronounced improvements in tasks requiring compositional or abstract reasoning where high-quality supervision signals are sparse but critical. This curriculum-aware reuse not only enhances data efficiency but also ensures that the model refines its behavior on the most informative examples during the final exploitation phase, directly contributing to ACPO’s leading average scores in both reasoning categories.

Furthermore, the advantage-aware adaptive clipping mechanism provides a fine-grained control over policy updates that standard PPO’s fixed $\epsilon$ cannot match. In complex multimodal settings, where token-level advantages vary significantly—e.g., a correct geometric deduction in Geo3k may yield high advantage, while a misaligned visual reference in Counting may produce low or negative advantage—ACPO dynamically widens the clipping bound for high-advantage tokens, enabling aggressive updates where the signal is strong, while constraining updates for ambiguous or noisy samples. This explains why ACPO achieves the best results on high-stakes benchmarks like We-Math (61.67\% at 3B, 69.15\% at 7B) and Geo3k (33.13\% at 3B, 41.58\% at 7B), where precise, confident reasoning steps are essential. The adaptive clipping thus acts as an implicit “reasoning confidence modulator,” aligning optimization intensity with the reliability of each learning signal—ultimately yielding a more robust, scalable, and consistently top-performing policy across diverse multimodal reasoning challenges.

\begin{table}[!t]
\centering
\caption{Comparative performance evaluation across vision-dependent and general tasks in 3B model sizes. \textbf{RED BOLD} indicates the best performance, and \underline{UNDERLINED} indicates the second-best performance.}
\label{tab:main_results}
\scalebox{0.62}{
\renewcommand{\arraystretch}{1.5} 
\begin{tabular}{l|c|
>{\columncolor[HTML]{FFFFFF}}c 
>{\columncolor[HTML]{FFFFFF}}c 
>{\columncolor[HTML]{FFFFFF}}c 
>{\columncolor[HTML]{FFFFFF}}c 
>{\columncolor[HTML]{FFFFFF}}l |
>{\columncolor[HTML]{FFFFFF}}c 
>{\columncolor[HTML]{FFFFFF}}c 
>{\columncolor[HTML]{FFFFFF}}c 
>{\columncolor[HTML]{FFFFFF}}l }
\hline
\multicolumn{1}{c|}{}                                 & {\cellcolor[HTML]{ECD9EC}\textbf{Overall}}                                                & \multicolumn{5}{c|}{\cellcolor[HTML]{D4EFFB}\textbf{Vision-Dependent Multimodal Reasoning}}                                                                                                                                                                           & \multicolumn{4}{c}{\cellcolor[HTML]{FFECD9}\textbf{General Multimodal Reasoning}}                                                                                                                                          \\ \cline{2-11} 
\multicolumn{1}{c|}{\multirow{-2}{*}{\textbf{Model}}} & \cellcolor[HTML]{FFFFFF}\textbf{Average(@8)}                    & \textbf{MathVerse-V}                    & \textbf{MMMU-Pro}                       & \textbf{Counting}                       & \multicolumn{1}{c|}{\cellcolor[HTML]{FFFFFF}\textbf{LogicVista}}                     & \multicolumn{1}{c|}{\cellcolor[HTML]{FFFFFF}AVG} & \textbf{MathVerse}                      & \textbf{Geo3k}                          & \multicolumn{1}{c|}{\cellcolor[HTML]{FFFFFF}\textbf{We-Math}}                        & \multicolumn{1}{c}{\cellcolor[HTML]{FFFFFF}AVG} \\ \hline
\cellcolor[HTML]{FFFFFF}DAPO-Off$_{\textbf{3B}}$               & \cellcolor[HTML]{FFFFFF}44.51\%                                 & 46.37\%                                 & 26.76\%                                 & 72.06\%                                 & \multicolumn{1}{c|}{\cellcolor[HTML]{FFFFFF}37.36\%}                                 & 45.63\%                                          & 49.72\%                                 & 24.60\%                                 & \multicolumn{1}{c|}{\cellcolor[HTML]{FFFFFF}54.70\%}                                 & 43.01\%                                         \\
\cellcolor[HTML]{FFFFFF}DAPO-On$_{\textbf{3B}}$               & \cellcolor[HTML]{FFFFFF}{\ul47.68\%}                                 & 48.32\%                                 & 28.69\%                                 & \cellcolor[HTML]{FFFFFF}{\color[HTML]{D83931} \textbf{73.88\%}}                                 & \multicolumn{1}{c|}{\cellcolor[HTML]{FFFFFF}39.07\%}                                 & {\ul 47.49\%}                                                    & 51.96\%                                 & {\ul 32.57\%}                                 & \multicolumn{1}{c|}{\cellcolor[HTML]{FFFFFF}{\ul59.30\%}}                                 & \cellcolor[HTML]{FFFFFF}{47.94}\%                                 \\
\cellcolor[HTML]{FFFFFF}PAPO$_{\textbf{3B}}$                          & \cellcolor[HTML]{FFFFFF}{47.26\%}                           & {\ul 49.04\%}                           & {\ul 29.31\%}                           & {65.88\%}                           & \multicolumn{1}{c|}{\cellcolor[HTML]{FFFFFF}{\color[HTML]{D83931} \textbf{41.41\%}}} & {46.41\%}                                    & {\ul 54.79\%}                           & {31.82\%}                           & \multicolumn{1}{c|}{\cellcolor[HTML]{FFFFFF}{ 58.57\%}}                           & {\ul 48.39\%}                                   \\
\cellcolor[HTML]{FFFFFF}ACPO$_{\textbf{3B}}$                          & \cellcolor[HTML]{FFFFFF}{\color[HTML]{D83931} \textbf{49.90\%}} & {\color[HTML]{D83931} \textbf{53.63\%}} & {\color[HTML]{D83931} \textbf{29.60\%}} & { \ul 72.75\%} & \multicolumn{1}{c|}{\cellcolor[HTML]{FFFFFF}{\ul 41.14\%}}                           & {\color[HTML]{D83931} \textbf{49.28\%}}          & {\color[HTML]{D83931} \textbf{57.41\%}} & {\color[HTML]{D83931} \textbf{33.13\%}} & \multicolumn{1}{c|}{\cellcolor[HTML]{FFFFFF}{\color[HTML]{D83931} \textbf{61.67\%}}} & {\color[HTML]{D83931} \textbf{50.74\%}}         \\
\hline
\end{tabular}
}

\end{table}

\begin{table}[!t]
\centering
\caption{Comparative performance evaluation across vision-dependent and general tasks in 7B model sizes. \textbf{RED BOLD} indicates the best performance, and \underline{UNDERLINED} indicates the second-best performance.}
\label{tab:main_results-2}
\scalebox{0.62}{
\renewcommand{\arraystretch}{1.5} 
\begin{tabular}{l|c|
>{\columncolor[HTML]{FFFFFF}}c 
>{\columncolor[HTML]{FFFFFF}}c 
>{\columncolor[HTML]{FFFFFF}}c 
>{\columncolor[HTML]{FFFFFF}}c 
>{\columncolor[HTML]{FFFFFF}}l |
>{\columncolor[HTML]{FFFFFF}}c 
>{\columncolor[HTML]{FFFFFF}}c 
>{\columncolor[HTML]{FFFFFF}}c 
>{\columncolor[HTML]{FFFFFF}}l }
\hline
\multicolumn{1}{c|}{}                                 & {\cellcolor[HTML]{ECD9EC}\textbf{Overall}}                                                & \multicolumn{5}{c|}{\cellcolor[HTML]{D4EFFB}\textbf{Vision-Dependent Multimodal Reasoning}}                                                                                                                                                                           & \multicolumn{4}{c}{\cellcolor[HTML]{FFECD9}\textbf{General Multimodal Reasoning}}                                                                                                                                          \\ \cline{2-11} 
\multicolumn{1}{c|}{\multirow{-2}{*}{\textbf{Model}}} & \cellcolor[HTML]{FFFFFF}\textbf{Average(@8)}                    & \textbf{MathVerse-V}                    & \textbf{MMMU-Pro}                       & \textbf{Counting}                       & \multicolumn{1}{c|}{\cellcolor[HTML]{FFFFFF}\textbf{LogicVista}}                     & \multicolumn{1}{c|}{\cellcolor[HTML]{FFFFFF}AVG} & \textbf{MathVerse}                      & \textbf{Geo3k}                          & \multicolumn{1}{c|}{\cellcolor[HTML]{FFFFFF}\textbf{We-Math}}                        & \multicolumn{1}{c}{\cellcolor[HTML]{FFFFFF}AVG} \\ \hline

\cellcolor[HTML]{FFFFFF}DAPO-Off$_{\textbf{7B}}$                          & 
\cellcolor[HTML]{FFFFFF}50.82\% & 
51.51\% & 
30.25\% & 
{\ul 89.25}\% & 
\multicolumn{1}{c|}{\cellcolor[HTML]{FFFFFF}38.17\%} & 
52.30\% & 
56.00\% & 
22.59\% & 
\multicolumn{1}{c|}{\cellcolor[HTML]{FFFFFF}53.94\%} & 
44.18\% \\

\cellcolor[HTML]{FFFFFF}DAPO-On$_{\textbf{7B}}$                          & 
\cellcolor[HTML]{FFFFFF}{56.05\%} & 
{57.51\%} & 
{35.20\%} & 
{87.19\%} & 
\multicolumn{1}{c|}{\cellcolor[HTML]{FFFFFF}{44.24\%}} & 
{56.04\%} & 
{61.62\%} & 
{32.51\%} & 
\multicolumn{1}{c|}{\cellcolor[HTML]{FFFFFF}{63.36\%}} & 
{ 52.50\%} \\

\cellcolor[HTML]{FFFFFF}PAPO$_{\textbf{7B}}$                          & 
\cellcolor[HTML]{FFFFFF} {\ul59.15\%} & 
{\ul 64.97\%} & 
{\ul 36.63\%} & 
{\color[HTML]{D83931} \textbf{89.81\%}} & 
\multicolumn{1}{c|}{\cellcolor[HTML]{FFFFFF} {\ul 46.07\%}} & 
{\color[HTML]{D83931} \textbf{59.37\%}} & 
{\color[HTML]{D83931} \textbf{69.53\%}} & 
{\ul40.25\%} & 
\multicolumn{1}{c|}{\cellcolor[HTML]{FFFFFF}{\ul66.79\%}} & 
{\ul58.85\%} \\

\cellcolor[HTML]{FFFFFF}ACPO$_{\textbf{7B}}$                          & 
\cellcolor[HTML]{FFFFFF}{\color[HTML]{D83931} \textbf{60.07\%}} & 
{\color[HTML]{D83931} \textbf{65.10\%}} & 
{\color[HTML]{D83931} \textbf{37.10\%}} & 
82.12\% & 
\multicolumn{1}{c|}{\cellcolor[HTML]{FFFFFF}{\color[HTML]{D83931} \textbf{47.93\%}}} & 
{{\ul58.06\%}} & 
{{\ul68.65\%}} & 
{\color[HTML]{D83931} \textbf{41.58\%}} & 
\multicolumn{1}{c|}{\cellcolor[HTML]{FFFFFF}{\color[HTML]{D83931} \textbf{69.15\%}}} & 
{\color[HTML]{D83931} \textbf{59.79\%}} \\

\hline
\end{tabular}
}

\end{table}

\begin{figure}[!h]
\centering
\includegraphics[width=\linewidth]{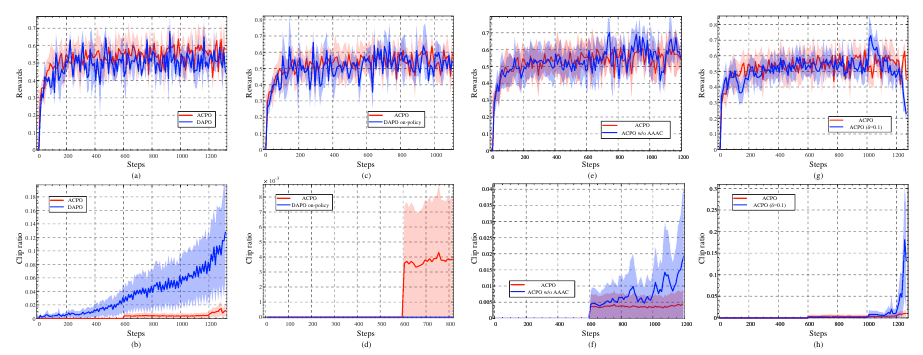}
\caption{Training Dynamics of Reward and Clip Ratio in Ablation and Baseline RL Experiments.}
\label{fig2}
\end{figure}

Fig.\ref{fig2} (a) and (c) show the cumulative reward curves of ACPO\textsubscript{3B} and the baseline DAPO\textsubscript{3B} under off-policy and on-policy settings, respectively. Both methods exhibit rapid initial performance improvement, indicating strong learning capability in the early training stages. However, under the off-policy setting, DAPO demonstrates significant reward fluctuations and achieves a notably lower convergence value compared to ACPO, suggesting an unstable policy update process that hinders long-term performance growth. In the on-policy setting, although DAPO reaches a convergence level close to that of ACPO, its reward trajectory remains highly volatile, indicating a lack of robustness in the optimization process. In contrast, ACPO consistently exhibits smoother convergence and higher final performance across both settings, highlighting its superior stability and generalization capability under different data collection strategies.

Fig.\ref{fig2} (b) and (d) present the corresponding clip ratio dynamics of both methods (for the 3B models). It can be observed that under the off-policy setting, the clip ratio of DAPO progressively increases during training, implying that a large portion of advantage signals are clipped. This restricts the magnitude of policy updates and prevents the model from fully leveraging the guidance of high-reward action directions. In contrast, ACPO maintains a consistently low clip ratio, indicating that it preserves more genuine advantage signals and allows more aggressive updates along high-advantage trajectories. This enables ACPO to explore higher-performance regions in the policy space, thereby achieving a superior performance upper bound. This behavior aligns with the superior convergence observed in Fig.\ref{fig2} (a), further demonstrating the effectiveness of ACPO’s policy update mechanism.

\subsubsection{Ablation Study}

\begin{table}[!h]
\centering
\caption{Ablation study of AAAC and scaling factor in 3B model size. \textbf{RED BOLD} indicates the best performance, and \underline{UNDERLINED} indicates the second-best performance.}
\label{tab:3}
\scalebox{0.60}{
\renewcommand{\arraystretch}{1.5} 
\begin{tabular}{l|c|
>{\columncolor[HTML]{FFFFFF}}c 
>{\columncolor[HTML]{FFFFFF}}c 
>{\columncolor[HTML]{FFFFFF}}c 
>{\columncolor[HTML]{FFFFFF}}c 
>{\columncolor[HTML]{FFFFFF}}l |
>{\columncolor[HTML]{FFFFFF}}c 
>{\columncolor[HTML]{FFFFFF}}c 
>{\columncolor[HTML]{FFFFFF}}c l}
\hline
\multicolumn{1}{c|}{}                                 & {\cellcolor[HTML]{ECD9EC}\textbf{Overall}}                                                & \multicolumn{5}{c|}{\cellcolor[HTML]{D4EFFB}\textbf{Vision-Dependent Multimodal Reasoning}}                                                                                                                                                                                     & \multicolumn{4}{c}{\cellcolor[HTML]{FFECD9}\textbf{General Multimodal Reasoning}}                                                                                                                                                          \\ \cline{2-11} 
\multicolumn{1}{c|}{\multirow{-2}{*}{\textbf{Model}}} & \cellcolor[HTML]{FFFFFF}\textbf{Average(@8)}                    & \textbf{MathVerse-V}                    & \textbf{MMMU-Pro}                       & \textbf{Counting}                       & \multicolumn{1}{c|}{\cellcolor[HTML]{FFFFFF}\textbf{LogicVista}}                     & \multicolumn{1}{c|}{\cellcolor[HTML]{FFFFFF}AVG}           & \textbf{MathVerse}                      & \textbf{Geo3k}                          & \multicolumn{1}{c|}{\cellcolor[HTML]{FFFFFF}\textbf{We-Math}}                        & \multicolumn{1}{c}{\cellcolor[HTML]{FFFFFF}AVG}                 \\ \hline
\cellcolor[HTML]{FFFFFF}ACPO w/o AAAC                   & 48.74\%                                                         & \cellcolor[HTML]{FFFFFF}{\ul 53.27\%}   & {\ul 28.91\%}                           & 68.88\%                           & \multicolumn{1}{c|}{\cellcolor[HTML]{FFFFFF}{\color[HTML]{D83931} \textbf{ 41.39\%}}}                           & \multicolumn{1}{c|}{\cellcolor[HTML]{FFFFFF}{\ul 48.11\%}} & \cellcolor[HTML]{FFFFFF}{\ul 56.36\%}   &  31.01\%                         & \multicolumn{1}{c|}{\cellcolor[HTML]{FFFFFF}{\ul 61.39\%}}                           & \multicolumn{1}{c}{{\ul 49.59\%}}                               \\ 
\cellcolor[HTML]{FFFFFF}ACPO$_{\delta = 0.10}$                  & 47.28\%                                                         & \cellcolor[HTML]{FFFFFF}{ \ul 52.38\%}   & {26.95\%}                           & 58.88\%                           & \multicolumn{1}{c|}{\cellcolor[HTML]{FFFFFF}{40.72\%}}                           & \multicolumn{1}{c|}{\cellcolor[HTML]{FFFFFF}{ 44.73\%}} & \cellcolor[HTML]{FFFFFF}{\ul 56.00\%}   &  {\color[HTML]{D83931} \textbf{33.78\%}}                        & \multicolumn{1}{c|}{\cellcolor[HTML]{FFFFFF}{\color[HTML]{D83931} \textbf{62.28\%}}}                           & \multicolumn{1}{c}{{\ul 50.69\%}}                               \\ 
\cellcolor[HTML]{FFFFFF}ACPO$_{\delta = 0.03}$                  & {\ul 47.52\%}                                                         & \cellcolor[HTML]{FFFFFF}{ 50.99\%}   & {\ul 27.60\%}                           & {\ul 68.19\%}                           & \multicolumn{1}{c|}{\cellcolor[HTML]{FFFFFF}{ \ul 40.97\%}}                           & \multicolumn{1}{c|}{\cellcolor[HTML]{FFFFFF}{\ul 46.94\%}} & \cellcolor[HTML]{FFFFFF}{54.24\%}   &  29.01\%                         & \multicolumn{1}{c|}{\cellcolor[HTML]{FFFFFF}{61.66\%}}                           & \multicolumn{1}{c}{{\ul 48.30\%}}                               \\ 
\cellcolor[HTML]{FFFFFF}ACPO$_{\delta = 0.05}$                        & \cellcolor[HTML]{FFFFFF}{\color[HTML]{D83931} \textbf{49.90\%}} & {\color[HTML]{D83931} \textbf{53.63\%}} & {\color[HTML]{D83931} \textbf{29.60\%}} & {\color[HTML]{D83931} \textbf{72.75\%}} & \multicolumn{1}{c|}{\cellcolor[HTML]{FFFFFF}{\color[HTML]{D83931} \textbf{ 41.14\%}}}                            & {\color[HTML]{D83931} \textbf{49.28\%}}                    & {\color[HTML]{D83931} \textbf{57.41\%}} & { \ul 33.13\%} & \multicolumn{1}{c|}{\cellcolor[HTML]{FFFFFF}{\ul 61.67\%}} & \cellcolor[HTML]{FFFFFF}{\color[HTML]{D83931} \textbf{50.74\%}} \\
\hline
\end{tabular}
}

\end{table}

As shown in Table ~\ref{tab:main_results} and ~\ref{tab:3}, we conduct an ablation study to evaluate the effectiveness of the AAAC mechanism. The results demonstrate that removing AAAC (i.e., ACPO w/o AAAC) leads to a performance degradation across multiple benchmark tasks, particularly in vision-dependent and general multimodal reasoning scenarios. Specifically, the overall accuracy drops from 49.90\% (ACPO) to 48.74\% (ACPO w/o AAAC), confirming that AAAC plays a crucial role in enhancing the model’s reasoning capability.

Fig.\ref{fig2} (e) and (f) present the ablation results after removing the AAAC module. It can be observed that after switching to the off-policy training setting, the model without AAAC exhibits a significant increase in the clipping ratio of advantage signals, accompanied by intensified fluctuations in the reward curve. This outcome is attributed to the curriculum learning mechanism, which continuously introduces more difficult samples during training. Without the AAAC module to effectively learn from such samples, the model fails to improve its performance when exposed to a large number of challenging instances; instead, it experiences a degradation in capability. These results fully demonstrate the importance of the AAAC mechanism in handling difficult samples, maintaining training stability, and enhancing overall performance.


In the ACPO algorithm, the clipping range of AAAC is set to 0.05. Fig.\ref{fig2} (g) and (h) present the experimental results when the AAAC range is expanded to 0.1. As shown in Fig.\ref{fig2} (g), during the early training phase (up to approximately 1000 steps), the reward curves under both settings are similar, indicating comparable learning behavior initially. However, beyond 1000 steps, the model with the larger AAAC range of 0.1 exhibits a noticeable decline in reward, demonstrating clear performance degradation. This behavior can be attributed to the overestimation of high-advantage signals caused by the excessively wide AAAC range, which results in policy updates that deviate too drastically from the reference policy. Such excessive deviation prevents the model from effectively learning useful policy information, ultimately leading to unstable or even divergent training. Further insights can be drawn from the clip ratio dynamics in Fig.\ref{fig2} (h). Although a larger AAAC range should theoretically allow more aggressive updates, a higher clip ratio is observed in practice. This indicates that when encountering difficult samples, the model fails to capture meaningful environmental feedback, still generating high advantage estimates that trigger more frequent clipping. This reflects instability in the policy optimization process. In conclusion, the configuration of the AAAC range significantly affects both training stability and learning efficiency. While intended to promote exploration, an excessively large range may lead to policy divergence and learning failure due to overly aggressive updates.

Furthermore, the ablation results in Tab.~\ref{tab:3} further validate the significant impact of the AAAC clipping range on the model’s final performance. When $\delta=0.10$ , the overly aggressive update strategy fails to improve performance and instead leads to training instability, hindering convergence. In contrast, when $\delta=0.03$ , the update rule becomes excessively conservative, limiting the model’s exploratory capacity and impeding effective learning. Through extensive empirical evaluation, we find that $\delta=0.05$ strikes an optimal balance between update magnitude and training stability, effectively trading off exploration and exploitation, and thereby achieving the best overall performance.
\section{Conclusion}
In this work, we present ACPO, a novel framework designed to overcome the limitations of static training schedules and fixed optimization boundaries inherent in prior reinforcement learning methods. The core innovation of ACPO lies in its dual adaptive mechanisms: a dynamic curriculum that orchestrates a smooth transition from stable exploration to efficient exploitation by intelligently scheduling data, and our proposed AAAC, which replaces the fixed clipping threshold with sample-wise dynamic bounds to enable more granular and effective policy updates. Extensive experiments validate the superiority of our approach: ACPO not only achieves a state-of-the-art average accuracy of 49.90\% across multiple complex multimodal reasoning benchmarks, significantly outperforming strong baselines like DAPO and PAPO, but also exhibits faster convergence and exceptional training stability. These advantages demonstrate that ACPO establishes a more efficient, robust, and adaptive optimization paradigm for the alignment of large-scale vision-language models.

\bibliographystyle{unsrt}  
\bibliography{references}

\newpage
\appendix
\section{Appendix}

In the initial training phase, the model is exposed primarily to "Easy" problems. These tasks are straightforward and involve direct application of basic concepts. For example, determining whether Addison can afford both items requires only simple arithmetic and comparison. Calculating the area ratio of similar triangles relies solely on a fundamental geometric property.

Through such elementary exercises, the model quickly learns core mathematical operations and basic geometric principles. This process establishes an essential knowledge foundation. It mirrors how students first learn arithmetic and simple geometry to build initial cognitive frameworks.

As training advances, the model encounters "Middle" difficulty problems. These require integrated application of knowledge and preliminary logical reasoning. For instance, finding a circle's radius given a perpendicular chord demands combining the perpendicular chord theorem with the Pythagorean theorem.

\begin{table}[!ht]
\scalebox{0.65}{
\begin{tabular}{l|l|l}
\hline
\textbf{Type} & \textbf{Image} & \textbf{Query} \\ \hline
Eazy          & \includegraphics[width=3cm,height=3cm,keepaspectratio]{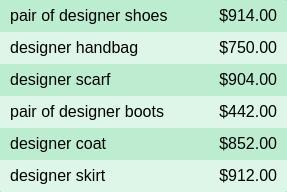} & With \$1,762.00, can Addison afford to purchase both a pair of designer shoes and a designer scarf? \\ \hline
Eazy          & \includegraphics[width=3cm,height=3cm,keepaspectratio]{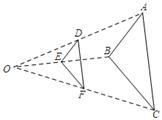}  & $\triangle ABC$ and $\triangle DEF$ are similar with point O as the center of similarity, and $\frac{OD}{OA}=\frac{1}{2}$. Then $\frac{S_{\triangle DEF}}{S_{\triangle ABC}}=?$ \\ \hline
Eazy          & \includegraphics[width=3cm,height=3cm,keepaspectratio]{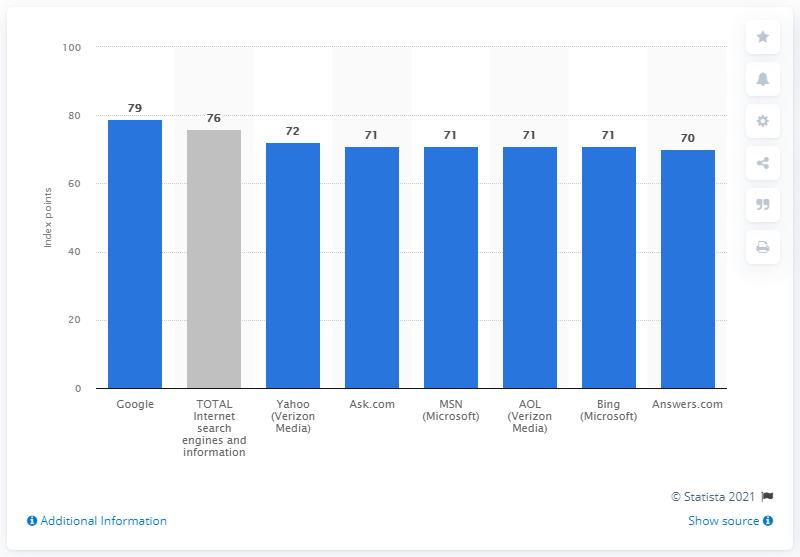} &  What was Google's ranking out of 100 ACSI index points in 2020? \\ \hline
Middle & 
\includegraphics[width=3cm,height=3cm,keepaspectratio]{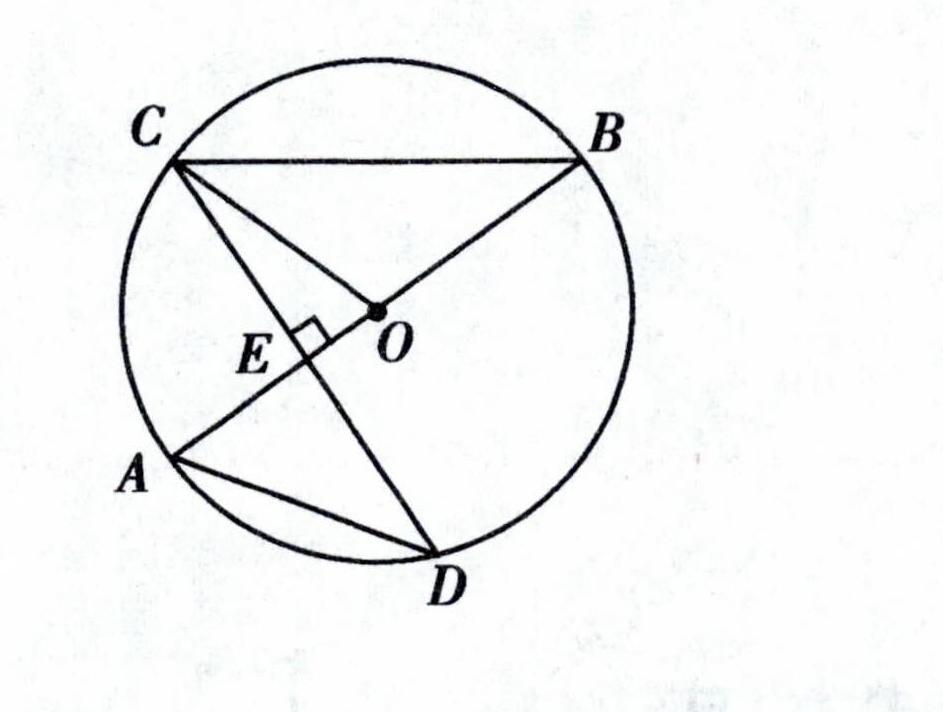} & 
$AB$ is the diameter, 
$CD$ is a chord, and 
$CD \perp AB$, $CD = 4\sqrt{2}$ and $AE = 2$, 
the radius is \_\_\_. \\ \hline
Middle        & \includegraphics[width=3cm,height=3cm,keepaspectratio]{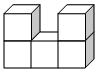}                &  A geometric solid is formed by 5 cubes with an edge length of 1. The surface area of this geometric solid is \_\_\_.           \\ \hline
Middle        &   \includegraphics[width=3cm,height=3cm,keepaspectratio]{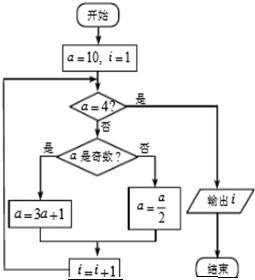}              &   In the following algorithm, the output value of i is  \_\_\_.          \\ \hline
Difficult     &  \includegraphics[width=3cm,height=3cm,keepaspectratio]{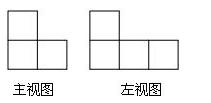}              &    The maximum number of small cubes that make up this geometric body is \_\_\_.            \\ \hline
Difficult     &    \includegraphics[width=3cm,height=3cm,keepaspectratio]{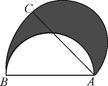}            &      $AB = 6$ is the diameter. The semicircle is rotated 45° clockwise around point A. What is the area of the shaded region?        \\ \hline
Difficult     &    \includegraphics[width=3cm,height=3cm,keepaspectratio]{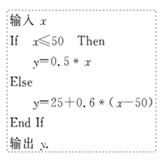}            &     According to the following algorithm statements, when the input $x$ is 60, the output value of $y$ is ?           \\ \hline
\end{tabular}
}
\end{table}

Another example is calculating the surface area of a composite solid. This requires careful spatial visualization and systematic counting of exposed faces. Algorithmic problems at this level involve tracing execution steps to deduce outputs.

At this stage, the model must synthesize previously acquired knowledge. It develops logical structuring skills and enhances comprehensive problem-solving abilities. This phase resembles students progressing to multi-concept exercises.

In the final training stage, the model faces "Difficult" problems. These feature high complexity and substantial cognitive challenges. For example, determining the maximum number of unit cubes in a solid requires advanced spatial reasoning and exploring various combinatorial possibilities.

Another challenging task involves calculating shaded areas after rotational transformations. This demands understanding geometric transformations and applying area decomposition strategies. Algorithmic problems may involve nested conditionals or iterative computations.

At this level, the model must deeply integrate all previously learned knowledge and skills. It demonstrates advanced capabilities like rigorous spatial imagination and multi-step logical deduction. This stage is analogous to students tackling contest-level problems requiring innovative application of concepts.

\end{document}